\pdfoutput=1

\documentclass[11pt]{article}

\usepackage[preprint]{acl}

\usepackage{times}
\usepackage{latexsym}

\usepackage[T1]{fontenc}

\usepackage[utf8]{inputenc}

\usepackage{microtype}

\usepackage{inconsolata}

\usepackage{graphicx}
\usepackage{pgfplotstable}
\usepackage{booktabs}
\usepackage{array}
\usepackage{multirow}
\usepackage{amsmath}
\usepackage{amsfonts}
\usepackage{float}
\usepackage{makecell}

\title{YIELD: A Large-Scale Dataset and Evaluation Framework for Information Elicitation Agents}

\author{Victor De Lima \and Grace Hui Yang  \\
         InfoSense Lab, Georgetown University\\
         \texttt{\{vad49, grace.yang\}@georgetown.edu}
         }

\begin{document}

\maketitle

\begin{abstract}
    Most conversational agents (CAs) are designed to satisfy user needs through user-driven interactions. However, many real-world settings, such as academic interviewing, judicial proceedings, and journalistic investigations, involve broader institutional decision-making processes and require agents that can elicit information from users. In this paper, we introduce Information Elicitation Agents (IEAs) in which the agent's goal is to elicit information from users to support the agent's institutional or task-oriented objectives. To enable systematic research on this setting, we present \textit{YIELD}, a 26M-token dataset of 2,281 ethically sourced, human-to-human dialogues. Moreover, we formalize information elicitation as a finite-horizon POMDP and propose novel metrics tailored to IEAs. Pilot experiments on multiple foundation LLMs show that training on YIELD improves their alignment with real elicitation behavior and findings are corroborated by human evaluation. We release YIELD under CC BY 4.0. The dataset, project code, evaluation tools, and fine-tuned model adapters are available at: \url{https://github.com/infosenselab/yield}.
    
\end{abstract}

\begin{figure}[t!]
    \centering
    \includegraphics[height=8.8cm]{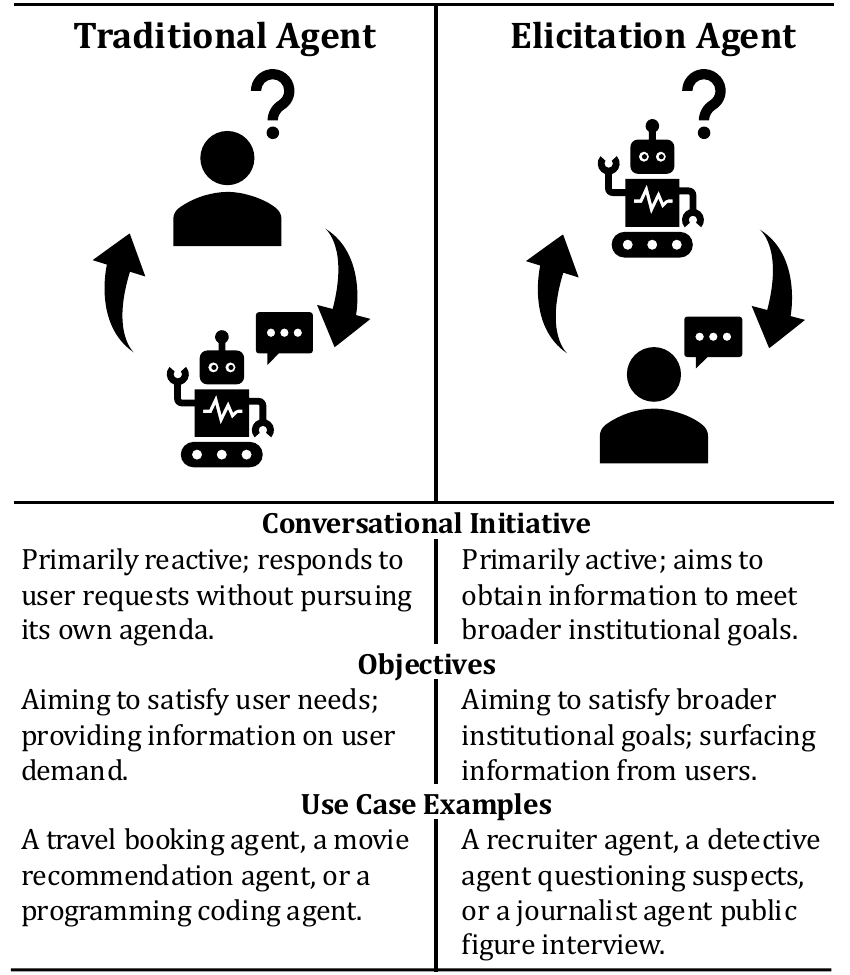}
    \vspace{-20pt}  
    \caption{Traditional vs. information elicitation agents.}\label{fig:pull_figure}
    \vspace{-20pt}
\end{figure}

\section{Introduction}
\label{sec:introduction}

Most conversational agents (CAs) are designed to satisfy user needs through user-driven interactions. However, many real-world settings, such as academic interviewing~\cite{taherzadeh_interview_2016}, judicial proceedings~\cite{supreme_court_of_the_united_states_moody_2024}, and journalistic investigations~\cite{van_susteren_voa_2018}, involve broader institutional decision-making processes and require agents that can elicit information from users. In this paper, we propose to study and support Information Elicitation Agents (IEAs), a new class of conversational agents designed to meet these goals. Figure~\ref{fig:pull_figure} illustrates the core distinctions between traditional CAs and IEAs across three dimensions: conversational initiatives, agent's objectives, and common use cases.  Table~\ref{tab:dialogue_types_examples} shows sample dialogues between human elicitors and human respondents in various domains, including oral history and judicial proceedings. The goal of this paper is to support the study and creation of agents that can behave like human elicitors. However, little research has been done for this purpose.

\begin{table*}[t!]
  \small
  \centering
    
  \begin{tabular}{p{7.6cm}  @{\hspace{0.3cm}}  p{7.6cm}}

  \toprule

  \textbf{Oral History} \cite{oglesby_oral_2009} & \textbf{Judicial Proceeding} \cite{supreme_court_of_the_united_states_moody_2024} \\

  \midrule

  
  \textbf{Elicitor}: "OK, then think back to when you went into the Peace Corps. What were you doing the year before you went in?" 

  \textbf{Respondent}: "I was in high school when I went in. I was one of the eighteen year olds that went into the Peace Corps."

  \textbf{Elicitor}: "Did you have any particular country or any place that you wanted to go?"

  \textbf{Respondent}: "South America always had a calling for me. I didn't have any particular country at all."
  & 


  \textbf{Elicitor}: "Is there any aspect of social media that you think is protected by the First Amendment?"

  \textbf{Respondent}: "Yes, Your Honor. I can certainly imagine platforms that would be subject to this law that indeed have First Amendment rights."

  \textbf{Elicitor}: "But why is it different?"

  \textbf{Respondent}: "Well -- well, there's a lot in there, Your Honor. Maybe I can start with the Hurley case."
  \\
  
  \toprule

  \textbf{Academic Interview} \cite{taherzadeh_interview_2016} & \textbf{Journalistic Investigation} \cite{van_susteren_voa_2018} \\

  \midrule


  \textbf{Elicitor}: "I was just wondering your opinion on whether biodiversity offsetting is likely to lead to more stringent or less stringent regulations around infrastructure projects?" 
  
  \textbf{Respondent}: "It's ill-informed people not understanding the system properly because at the end of the day biodiversity offsetting doesn't affect any existing protections."
  
  \textbf{Elicitor}: "What's been the reception of businesses to biodiversity offsetting extending the mitigation hierarchy?"
  
  \textbf{Respondent}: "Well, the reception that I've had when I've been speaking to them has been really quite positive." 
  &


  \textbf{Elicitor}: "This is your second trip to Mexico, but why are you here now in Mexico?"

  \textbf{Respondent}: "So, as the new government makes this transition beginning on Dec. 1, we’re working diligently to make sure we have a solid foreign policy relationship with them."
  
  \textbf{Elicitor}: "That caravan that is coming up from Guatemala heading to Mexico and then presumably maybe some place else, maybe the United States: What’s Mexico say about that?"
  
  \textbf{Respondent}: "So Mexico agrees that it is not constructive to have unlawful migration transit from these Northern Triangle countries through Mexico into the United States."
  \\
  
  \bottomrule
  \end{tabular}
  \vspace{-5pt}
  \caption{Shortened examples of information elicitation dialogues. Sources are included next to domain names.}\label{tab:dialogue_types_examples}
  \vspace{-15pt}
\end{table*}

IEAs differ fundamentally from other conversational agents in both whose goals they serve and how success is defined. Conversational agents are generally designed to help users accomplish user-initiated tasks, with the user controlling the agenda and direction of the interaction. IEAs, by contrast, act on behalf of an external institutional objective and use conversation as a means to extract information from the respondent. Crucially, at any point in the dialogue there is not a single optimal question or expected answer; instead, there are many possible directions, each potentially yielding valuable information. The agent’s task is therefore to choose questions that are likely to surface informative and relevant facts, maximizing the cumulative value of what is uncovered rather than converging on a single correct outcome.

The task of creating IEAs is challenging for several reasons. First, the agent must act in an extremely high-dimensional action space, where every utterance can shift the trajectory of the dialogue. Second, defining what makes information ``informative'' is inherently nuanced and context-dependent, bringing subjectivity into both training and evaluation. These difficulties are further amplified in long dialogues, where agents must maintain topic coherence over many turns while responding adaptively to unpredictable or even adversarial user replies. Additionally, the amount of publicly available data for training IEAs is limited. To the best of our knowledge, no existing dataset is available to support the development of IEAs. In this paper, we make the following contributions:

\begin{table*}[t!]
    \small
    \centering
    
    \begin{tabular}{
      >{\centering\arraybackslash}m{1.6cm}
      >{\centering\arraybackslash}m{1.4cm}
      >{\raggedleft\arraybackslash}m{1.5cm}
      >{\raggedleft\arraybackslash}m{1.7cm}
      >{\raggedleft\arraybackslash}m{1.7cm}
      >{\raggedleft\arraybackslash}m{1.4cm}
      >{\raggedleft\arraybackslash}m{1.4cm}
    }
    
    \toprule
      
    Dataset & \# Domains & \# Dialogues & Avg. Turns Per Dialogue & Avg. Tokens Per Turn &  \# Turns & \# Tokens \\
    \midrule
    DSTC2     & 1   & 1,612   & 14.49   & 8.54   & 23,354    & 199,431   \\
    MultiWOZ  & 7   & 8,438   & 13.46   & 13.13  & 113,556   & 1,490,615 \\
    SGD       & 16  & 16,142  & 20.44   & 9.75   & 329,964   & 3,217,149 \\
    \textbf{YIELD} & 4   & 2,281 & \textbf{171.07} & \textbf{67.27} & 390,205 & \textbf{26,249,014} \\

  \bottomrule
  \end{tabular}
  \vspace{-5pt}
  \caption{Commonly used dialogue dataset statistics.}\label{tab:comparison_table}
  \vspace{-10pt}
\end{table*}

\begin{itemize}
    \vspace{-5pt}
    \item We introduce IEAs as a distinct conversational paradigm and release \textit{YIELD: Information-Elicitation Learning Dialogues} (YIELD), the first large-scale, ethically sourced IEA dataset. YIELD consists of real, human-to-human, English information-elicitation dialogues sourced exclusively from repositories with clear public domain or Creative Commons licenses. YIELD contains four domains and is larger than existing task-oriented datasets (See Table \ref{tab:comparison_table}). We publish both the standard, full dialogue version of the dataset and a rolling-window segmented version used for the experiments in this study.
    \vspace{-7pt}
   \item We propose and operationalize a few IEA-specific evaluation metrics: \textit{Conformity}, \textit{Progression}, and \textit{Turn-Length Ratio}.
   \vspace{-7pt}
   \item We further formalize information elicitation as a finite-horizon \textit{Partially Observable Markov Decision Process} (POMDP) and use offline reinforcement learning (ORL) with a minimal proxy reward to show that dialogue-level optimization with YIELD is competitive with supervised fine-tuning (SFT).
   \vspace{-7pt}
   \item Pilot experiments across multiple foundation LLMs show that fine-tuning on YIELD improves distributional and behavioral alignment with human elicitation dialogues. These findings are confirmed by human evaluations.
   \vspace{-7pt}
\end{itemize}

The remainder of the paper is organized as follows. Section \ref{sec:related} reviews related work. Section \ref{sec:construction_process} describes the YIELD dataset. Section \ref{sec:formulation_and_implementation} introduces the formalization of the information elicitation problem. Section \ref{sec:metrics} presents IEA evaluation metrics. Section \ref{sec:experiment} details the pilot experiments and human evaluation; and Section \ref{sec:conclusion} concludes the paper.

\vspace{-5pt}
\section{Related Work}
\label{sec:related}
\vspace{-5pt}

Dialogue datasets are instrumental in advancing research in conversational agents. Annotated corpora were central to the development of CAs. Early work like the ATIS corpus used labeled intents to drive system research \cite{hemphill_atis_1990}, while more recent ones, such as the Dialog State Tracking Challenge (DSTC) series \cite{henderson_third_2014,williams_dialog_2013}, shifted attention toward dialogue state tracking. This emphasis on interaction laid the groundwork for exploring multi-turn information-seeking, where systems make progress through user feedback rather than executing static plans \cite{tang_reinforcement_2017}. More recent datasets, such as Schema-Guided Dialogue (SGD)\cite{rastogi_towards_2019} and MultiWOZ \cite{budzianowski_multiwoz_2018}, expanded in size and domain coverage, with 16,142 and 8,438 dialogues, respectively. Nonetheless, these existing datasets typically have short dialogues. For instance, DSTC2, SGD, and MultiWOZ averaged 14.49, 20.44, and 13.46 turns per dialogue, respectively. In contrast, the proposed YIELD dataset offers substantially longer dialogues, averaging 171 turns per dialogue. Table~\ref{tab:comparison_table} provides a comparison among these popular dialogue datasets and YIELD. 

There is growing interest in role-aware CAs,  which adopt defined personas and pursue structured goals. These agents have gained popularity in a range of applications, from mental health support \cite{fitzpatrick_delivering_2017} to legal assistance \cite{thomson_reuters_casetext_2025}. Common development approaches use foundation LLMs fine-tuned with domain-specific datasets, often combined with reinforcement learning and human feedback to maintain role consistency and behavioral alignment \cite{kong_better_2023}. In parallel, research on proactive CAs explores how conversational systems can drive interactions forward, such as initiating questions, shifting topics, and helping users make decisions \cite{deng_towards_2024,liao_proactive_2023}. These systems have been used in exploratory search and evaluative tasks such as LLM-as-an-Interviewer \cite{kim_llm-as--interviewer_2024} and LLM-as-a-Judge \cite{zheng_judging_2023}. Both role-aware and proactive agents benefit from curated dialogue datasets tailored to their specific domains and objectives.

The Information Elicitation Agents (IEAs) presented in this paper represent a natural extension of these trends. IEAs obtain information from human respondents to support the agents' information needs, such as information needs to serve for institutional goals. Yet, despite increasing interest in these capabilities, no existing dataset has been designed to support IEAs. To our knowledge, YIELD is the first large-scale, ethically sourced benchmark created to enable the training and evaluation of IEAs across multiple domains.

\vspace{-5pt}
\section{Dataset Construction}
\label{sec:construction_process}

Our dataset construction process consists of three stages: data acquisition, annotation, and dataset standardization. 

\subsection{Data Acquisition}

We chose the data sources with the following aims: (1) to maximize coverage of elicitation settings and (2) to ensure legal usability through public domain or Creative Commons licensing. Many of the files, particularly PDF and DOCX transcripts, to the best of our knowledge, had not previously been processed into any dialogue dataset. We categorized the identified sources into four domains of elicitation dialogues. For a full description of the data sources, see Appendix~\ref{appendix:data_sources}.

\subsection{Annotation}

The collected transcripts appeared in various formats, including HTML, TXT, PDF, and DOCX. Unifying these different formats and heterogeneous materials required extensive pre-processing. Automated text extractors regularly introduced errors during conversion and sources contained individual issues such as speaker names that appeared without distinguishing markers and often varied in form within the same document. These and other issues made automatic parsing and tagging unreliable. Therefore, we manually corrected and annotated the transcripts over several weeks, removing extraneous text, standardizing speaker references, and marking turns and speaker roles. Further details on the annotation and preprocessing pipeline are provided in the dataset documentation.

\vspace{-5pt}
\subsection{Dataset Standardization}
\label{sec:data_standardization}

We then passed the manually tagged files through standardization scripts to enforce a uniform schema (Table \ref{tab:base_schema}). The resulting dataset includes 148 dialogues from the Academic Interviews domain, 129 Journalistic Investigations, 621 Judicial Dialogue, and 1,383 Oral History; with a total of 2,281 dialogues, 390,205 turns, and 26,249,014 tokens. Table~\ref{tab:aggregated_domain_df} shows a breakdown of YIELD’s dataset statistics by domain and source. We split the dialogues into training, development, and test sets using stratified sampling for each domain. The development and test sets each contain 10\% of the data (228 and 229 dialogues, respectively), while the training set contains the remaining 80\% (1,824). We grouped the dialogue files into sets of 128 dialogues each. The last file in a set may contain fewer than 128 dialogues. The unzipped size of YIELD is 201.06 MB. All dialogues are organized around two roles (elicitor and respondent), but may include multiple speakers within a role (e.g., group interviews). In this work, all implementations are done at the role level, while the released dataset preserves speaker identities to enable future multi-speaker modeling.

\section{IEA Formulation and Implementation}
\label{sec:formulation_and_implementation}

We formalize information elicitation as a finite-horizon POMDP and use offline reinforcement learning with Advantage-Weighted Regression (AWR) \cite{peng_advantage-weighted_2019} to fine-tune the model.

\begin{table}[t!]
    \small
    \centering    
    \begin{tabular}{m{2cm}m{3.6cm}}
      \toprule
      \textbf{Field} & \textbf{Description} \\
      \midrule
      dialogue\_id & Unique dialogue identifier. \\ 
      metadata & Metadata varies by collection. \\ 
      broad\_source & Collection's primary archive. \\ 
      domain & General content category. \\ 
      title & Dialogue name (if available). \\ 
      elicitors & Speaker's names/identifiers. \\ 
      respondents & Speaker's names/identifiers. \\ 
      languages & Dialogue languages. \\ 
      turns & Conversation turns sequence. \\ 
      turns.turn\_id & Unique turn identifier . \\ 
      turns.timestamp & Turn timestamp, if any. \\ 
      turns.speaker & Speaker name/identifier. \\ 
      turns.role & "elicitor" or "respondent". \\ 
      turns.utterance & The turn's spoken content. \\ \bottomrule
    \end{tabular}
    \caption{Dataset schema.}\label{tab:base_schema} 
    \vspace{-15pt}
\end{table}

\subsection{States, Actions, and Observations}

We model the process of elicitation as a sequence of interactions between an \textit{elicitor} (interviewer) and a \textit{respondent} (interviewee), where the elicitor is the agent and the respondent is part of the environment. The respondent possesses information that is unobservable by the elicitor, such as knowledge, opinions, or preferences, that the elicitor aims to infer from conversational interaction. This setup aligns naturally with the framework of a POMDP. 

At each turn, the environment has a latent state $X_t \in \mathcal{X}$ that represents the respondent’s hidden information. The elicitor takes an action $A_t \in \mathcal{A}$ in the form of a natural language utterance, after which the environment responds with a respondent utterance (observation) $O_{t+1} \in \mathcal{O}$ that depends on both the latent state $X_t$ and the elicitor's action $A_t$, leading to the next elicitor utterance $A_{t+1}$. This dynamic continues until the conversation is over. The \textit{conversation history} at time $t$ is a sequence of elicitor and respondent utterances\begin{equation}H_t = (A_{t-w}, O_{t-w+1}, \ldots, A_{t-1}, O_t),\end{equation}where $w$ is the window size in turns that is available to the agent at time $t$ when producing its next utterance. When the conversation history includes the elicitor's latest utterance $A_t$, we denote it $H^s_t$.

In the traditional POMDP formulation, the agent maintains a belief state $S_t$, which is a probability distribution over possible hidden environment states $X_t$. This distribution encodes the agent’s available information about the hidden variables governing the environment at time $t$ and serves as the agent's basis for decision making. In the dialogue setting, however, explicitly maintaining such a distribution is not feasible as $X_t$ represents unbounded and continuously changing information (everything the respondent might know, think, or believe). To make the problem tractable, we represent the information available to the elicitor at time $t$ with token-level hidden representations $S_t = f_\theta(H^s_t)$ produced by a causal language model, where $f_\theta(H^s_t)$ denotes the model’s forward pass over the full segment $H^s_t$ parameterized by $\theta$. Because the model is autoregressive, each token’s representation depends only on the tokens preceding it, meaning that the hidden states corresponding to the context $H_t$ encode everything the model has ``seen'' in the current turn’s context. $S_t$ provides the sufficient context from which the likelihood of $A_t$ can be evaluated, functionally mirroring the role of the belief state in the traditional POMDP formulation.

\begin{table}[t!]
\small
\centering
\renewcommand{\arraystretch}{1.395}
\begin{tabular}{
  >{\centering\arraybackslash}m{1.4cm}
  >{\centering\arraybackslash}m{1.0cm}
  >{\raggedleft\arraybackslash}m{1.0cm}
  >{\raggedleft\arraybackslash}m{1.1cm}
  >{\raggedleft\arraybackslash}m{1.1cm}
}

\toprule

Domain & Broad Source & \makecell{Dia-\\logues} & Turns ($\times10^3$) & Tokens ($\times10^3$) \\
\midrule

Academic      & Harvard   & 136  & 26.3  & 654.4  \\
Academic      & Other     & 12   & 1.2   & 56.7   \\
Journalistic  & VOA       & 73   & 2.2   & 150.2  \\
Journalistic  & Wikinews  & 56   & 3.2   & 172.5  \\
Judicial      & Oyez      & 621  & 162.9 & 7,655.5 \\
Oral          & JFK       & 270  & 83.0  & 3,056.4 \\
Oral          & JSC       & 1,012 & 102.0 & 13,626.4 \\
Oral          & NARA      & 101  & 9.5   & 876.9  \\

\midrule

Total &  & 2,281 & 390.2 & 26,249.0 \\

\bottomrule
\end{tabular}

\caption{Domain and source statistics.}
\label{tab:aggregated_domain_df}
\vspace{-10pt}
\end{table}

\subsection{Reward Signals}
\label{sec:reward_signals}

Each elicitor action $A_t$ is associated with a scalar reward $R_{t+1} = r(H_t, A_t, O_{t+1})$ reflecting the informational value of the respondent’s subsequent response $O_{t+1}$. We convert the reward signals into return-to-go format, in which $R_{t+1}$ at each elicitor turn is computed in reverse temporal order as the discounted sum of future rewards, $R^{\text{tg}}_t = \sum_{k=0}^{T-t} \gamma^k R_{t+k+1}$, where $\gamma$ is a discount factor. Unlike supervised fine-tuning, incorporating returns-to-go allows the learning process to take into account the effects of the agent's utterance on the entire remaining dialogue, even if the agent only sees a dialogue window at each time step.

\vspace{-6pt}
\subsection{Value Function and Advantage Estimation}
\label{sec:advantage_estimation}

We follow Advantage-Weighted Regression (AWR) \citep{peng_advantage-weighted_2019} to train the elicitor policy in an offline setting to bias learning toward elicitor utterances that yield higher long-term informational return. AWR estimates a state-value function $v_\psi(S_t)$ and computes advantages as the difference between empirical returns-to-go and the value prediction. In our implementation, the value function is realized as a linear value head on top of the LLM’s hidden representations. Because dialogue windows are padded to a fixed length, we use the hidden state of the end-of-sequence token following the elicitor’s utterance $A_t$ as input to the value head instead of the final hidden state.

\vspace{-5pt}
\subsection{Policy and Policy Optimization}
\label{sec:policy_optimization}

The elicitor’s policy $\pi_\theta$ is implemented by the same causal language model that produces the state representations $S_t = f_\theta(H^s_t)$. The policy defines an autoregressive distribution over token sequences representing the elicitor’s next utterance $A_t$, conditioned on the preceding dialogue context: $
\pi_\theta(A_t \mid H_t)
= \prod_{\ell=1}^{L}
\pi_\theta(y_\ell \mid H_t, y_{<\ell})$, where $A_t = (y_1, \ldots, y_L)$ denotes the sequence of output tokens for the elicitor’s turn.

During policy optimization, we aim to maximize the likelihood the model assigns to the offline dataset's utterances $A_t$. We employ parameter-efficient fine-tuning (PEFT) using Low-Rank Adaptation (LoRA) adapters \cite{hu_lora_2021} to enable efficient learning, in which the base causal language model remains frozen, and trainable low-rank matrices are inserted into its attention projection layers. During AWR optimization, only these adapter parameters and the value head receive gradient updates. The model processes the full segment $H^s_t = (H_t, A_t)$, predicting token probabilities autoregressively. Then, we mask and remove context and padding tokens so that gradients consider only the $A_t$. The weighted policy loss is: 
\begin{equation}\mathcal{L}_\pi(\theta)
= - \frac{1}{|\mathcal{B}|} \sum_{i \in \mathcal{B}}
\bar{w}_i
\log \pi_\theta(A_i \mid S_i),\end{equation}
where each $\bar{w}_i$ scales the contribution of sample $i$ within the mini-batch $\mathcal{B}$ according to its estimated advantage, departing from standard supervised fine-tuning where all examples contribute equally regardless of their quality.

The critic (value head) is trained jointly with the policy network to minimize the mean-squared error between predicted and empirical returns:
\begin{equation}\mathcal{L}_v(\psi)
= \frac{1}{|\mathcal{B}|} \sum_{i \in \mathcal{B}}
\big(v_\psi(S_i) - R^{\text{tg}}_i\big)^2.
\end{equation}

The combined optimization objective is:
\begin{equation}
\mathcal{L}(\theta, \psi)
= \mathcal{L}_\pi(\theta) + \mathcal{L}_v(\psi),\end{equation}
which jointly updates the actor (policy) and critic (value head).

\section{Evaluation for Information Elicitation Agents}
\label{sec:metrics}

In this section, we propose a set of key dimensions for evaluating IEAs. These dimensions go beyond conventional dialogue evaluation approaches, such as BLEU~\cite{papineni_bleu_2001}, task success~\cite{walker_paradise_1997}, or response appropriateness~\cite{see_what_2019}, by focusing on the goals and dynamics of information elicitation. We release an open-source Python package for computing all proposed evaluation metrics (included in the project repository).

\subsection{Conformity}
\label{sec:conformity}

\textit{Conformity} measures whether an IEA agent could generate dialogue output that follows the distributional patterns typical of real human elicitation dialogues. We identify two dimensions that are particularly informative as to whether the agent acquired the data's characteristics: (1) \textit{perplexity} computed by evaluating real human utterances from the testing set under the trained IEA model and (2) \textit{Response length}, measured as a simple token count. Perplexity reflects how well the model captures the linguistic content of human dialogue, while utterance length captures a critical stylistic property that varies significantly across large language models and strongly influences conversational behavior.

For perplexity, we condition the models on the dialogue history and score the \textit{real} human elicitor utterance for each block in the test set. We restrict the loss calculation to elicitor tokens so that context tokens do not affect the score and compute perplexity in the standard manner as the exponential of the token-level cross-entropy: $\text{PPL} = \exp\!\left( -\frac{1}{N} \sum_{t=1}^N \log p_\theta(y_t \mid y_{<t}, x) \right)$, where $x$ denotes the dialogue context and $y_t$ denotes the $t$-th token of the elicitor’s reference utterance. We report micro perplexity, which consists of pooling across all target tokens before exponentiation. Response length is measured as the number of tokens produced by the model’s tokenizer.

\subsection{Progression}

\textit{Progression} (P) measures the extent to which a conversation moves forward rather than stagnating on the same topic. For IEAs, this is a crucial behavioral indicator that an effective agent should guide the dialogue toward new, relevant information, rather than looping over already-covered ground. Traditional metrics often focus on utterance coherence or intent-matching, but these do not account for \textit{forward momentum} in the interaction. We quantify Progression as the average (decayed) cosine distance between each utterance embedding and a window of preceding utterance embeddings:
\begin{equation}P(U) = \frac{1}{L-k} \sum_{t=k+1}^{L}\frac{\sum_{j=1}^{k} \gamma^j \left(1 - \frac{u_{t-j} \cdot u_{t}}{\Vert u_{t-j}\Vert  \Vert u_{t}\Vert}\right)}{\sum_{j=1}^{k} \gamma^j},\end{equation}
where $U=(u_1,...,u_L)$ is the sequence of embeddings of the dialogue's utterances, $u_t$ is the embedding of the $t$-th utterance, and $L$ is the total number of utterances. $k$ is the context window size and $\gamma \in (0,1]$ is a decay factor that reduces the importance of older utterances. Parameters $k>1$ or $\gamma<1$ give a direction to the metric by introducing asymmetry. Setting $k=1$ (the conversation becomes a chain of pairs) and $\gamma=1$ (no decay) can make the formula symmetric: $P(U) = \frac{1}{L-1} \sum_{t=1}^{L-1} \left(1 - \frac{u_t \cdot u_{t+1}}{\Vert u_t\Vert \, \Vert u_{t+1}\Vert}\right)$. Since cosine distance lies in $[0, 2]$, so does $P$.

\subsection{Turn-Length Ratio}

\textit{Turn-Length Ratio} (TLR) compares the average verbosity of the respondent against that of the elicitor. In effective elicitation settings, the agent should speak concisely and strategically to draw out longer, content-rich responses from the user. This asymmetry suggests a successful elicitation dynamic. The ratio is defined as:
\begin{equation}
TLR(T_r, T_e) = \frac{\tfrac{1}{n_r}\sum_{i=1}^{n_r} T_{r,i}}{\tfrac{1}{n_e}\sum_{i=1}^{n_e} T_{e,i}},
\end{equation}
where $T_{r}=(T_{r,1},...,T_{r,n_{r}})$ is the sequence of token counts for respondent turns, $T_{e}=(T_{e,1},...,T_{e,n_{e}})$ is the sequence of token counts for elicitor turns, and $n_r$, $n_e$ are their respective turn counts. A ratio $TLR > 1$ indicates that respondents are producing more verbose turns than elicitors.

\section{Pilot Experiments}
\label{sec:experiment}

We conduct pilot experiments to evaluate model-generated elicitor utterances conditioned on fixed dialogue states and assess policy alignment with expert behavior without introducing live respondents, establishing necessary precursors to interactive evaluation. We selected the (1) Meta AI Llama-3.1-8B-Instruct \cite{meta_llama_llama-31-8b-instruct_2024} as our primary model due to computational considerations and its open-source status. To explore performance across related open-source variants, we also employed (2) DeepSeek-R1-Distill-Llama-8B \cite{deepseek-ai_deepseek-r1_2025}, a distilled variant of comparable size, and (3) Llama-3.2-3B-Instruct \cite{meta_llama_llama-32-3b-instruct_2024}, a smaller-scale variant.

\subsection{ORL Models}

The agents trained with offline reinforcement learning (ORL) follow the formulation in Section \ref{sec:formulation_and_implementation}. All experiments were executed using the \texttt{accelerate} framework for distributed training across three NVIDIA RTX A6000 GPUs. We used $T = 1$ and $\alpha = 0.25$ for computing the AWR weights for all models. Models were trained using YIELD's training split formatted as described in Section~\ref{sec:experimental_dataset}. We used the \textit{PEFT} library \cite{mangrulkar_peft_2022} to configure and apply LoRA.

\begin{table*}[t!]
    \small
    \centering
    \begin{tabular}{
      >{\centering\arraybackslash}m{1.2cm}
      >{\centering\arraybackslash}m{1.2cm}
      >{\centering\arraybackslash}m{1.2cm}
      >{\centering\arraybackslash}m{1.2cm}
      >{\centering\arraybackslash}m{1.2cm}
      >{\centering\arraybackslash}m{1.2cm}
      >{\centering\arraybackslash}m{1.2cm}
      >{\centering\arraybackslash}m{1.2cm}
      >{\centering\arraybackslash}m{1.2cm}
    }
    
    \toprule
      
    Domain          & Segments          & Tokens               & Tokens / Segment  & Token Range     & Reward Avg    & Reward Range    & RTG Avg       & RTG Range         \\
    \midrule
    Academic        &  11,036           & 2,255,573            & 204.4             & 81-512          & 0.3	         & 0-29	           & 2.5	       & 0-36.5            \\
    Journalistic    &  1,331            & 414,870 	           & 311.7             & 99-511          & 1.3	         & 0-26	           & 8.7	       & 0-117.1           \\
    Judicial        &  48,679           & 14,920,246           & 306.5             & 91-512          & 0.5	         & 0-18	           & 4.7	       & 0-40.4            \\
    Oral            &  42,408           & 12,156,466           & 286.7             & 83-512          & 1.6	         & 0-223           & 13.6	       & 0-453.0           \\

    \midrule

    \textbf{Total}  & \textbf{103,454}  & \textbf{29,747,155}  & \textbf{277.3}     & \textbf{81-512}  & \textbf{0.9}  & \textbf{0-223}  & \textbf{7.4}  & \textbf{0-453.0} \\

  \bottomrule

  \end{tabular}
    \caption{Experimental YIELD dataset statistics. Reward outliers in Oral History are due to prompts that led to the respondent sharing stories of significant length.}\label{tab:experimental_dataset_statistics}

\end{table*}

\subsection{Reward Signal}

As a first-step reward formulation to validate the proposed elicitation framework, we operationalize rewards $r$ using a factual novelty signal as proxy for informational gain, measuring how much previously unmentioned content an elicitor action surfaces. We apply a spaCy-based NER model \cite{honnibal_spacy_2020} to each utterance and maintain a cumulative set of previously mentioned entities, events, or facts. After the elicitor’s turn, the respondent’s reply is scanned for entities not yet observed; the elicitor is rewarded in proportion to each newly introduced entity. Formally, the reward is $R_{t+1} = \big|\mathcal{E}_{t+1} \setminus \mathcal{E}_{\le t}\big|$, where $\mathcal{E}_{t+1}$ denotes the set of entities extracted from the respondent’s response, $\mathcal{E}_{\le t}$ the accumulated set from all prior turns, and $\setminus$ denotes the set difference operator. To mitigate potential reward-gaming behaviors, we compute factual novelty under several constraints. Entities produced by the elicitor are also added to the cumulative entity set, but they do not contribute reward, so that entities subsequently repeated by the respondent are not misattributed. Repeated entities yield no additional credit, therefore, agents that allow conversations to linger on the same content accrue fewer reward. While entity count alone could be gamed in unconstrained settings (e.g., a model providing a list of entities as utterance), this risk is mitigated here because rewards are computed on curated, real human–human dialogues.

\subsection{Experimental Dataset}
\label{sec:experimental_dataset}

We adapted YIELD to fit the experimental setup by segmenting the dialogues into fixed length \textit{blocks} using a sliding window of six turns. Each block ends with an elicitor utterance, with the preceding five turns serving as conversational context. We added a base system instruction to each block (``\texttt{Act as an information elicitation agent for \{domain\}.}''). We removed blocks exceeding 512 tokens to maintain computational feasibility and consistent utterance length. We removed segments ending with an elicitor utterance of two words or fewer as we found these generally impacted training negatively by oversampling \textit{backchannel} statements (e.g., ``Okay'', ``Hmm'', ``Got it''). Then, we computed rewards and returns-to-go with a $\gamma=0.9$ discount factor on the full dialogues and annotated them. There are a total of 103,454 segments with 83,181 on the training set and 9,988 and 10,285 on the development and testing sets, respectively.  Descriptive statistics are available in Table~\ref{tab:experimental_dataset_statistics}, while a sample line is available in Appendix~\ref{appendix:experimental_dataset_sample}.

\subsection{Prompted and SFT Models}

We compare against two versions of the foundational models: (1) prompt-only and (2) supervised fine-tuning (SFT). Prompt-only models use the same base system instruction as ORL and SFT, plus an additional model-agnostic prompt (Appendix~\ref{appendix:baseline_prompt}). For SFT, we apply PEFT with LoRA adapters using the \textit{SFTTrainer} \cite{von_werra_trl_2020}.

\begin{table}[t!]

  \small

  \centering

  \begin{tabular}{
    m{2.8cm}
    >{\centering\arraybackslash}m{0.7cm}
    >{\centering\arraybackslash}m{0.7cm}
    >{\centering\arraybackslash}m{0.7cm}
    >{\centering\arraybackslash}m{0.7cm}
  }
  
  \toprule

   Model & Ac	& Jo	& Ju	& Or	 \\

  \midrule

    \multicolumn{5}{c}{\textit{Micro Perplexity (models vs real utterances)}} \\

  Llama-3.1-8B Base & 46.9 & 29.3 & 22.6 & 30.4 \\
  Llama-3.1-8B SFT    & 10.9 & 13.0 & 10.9 & 11.0 \\
  Llama-3.1-8B ORL    & 12.5 & 13.2 & 11.3 & 11.6 \\

  \midrule

  DeepSeek-R1 Base & 78.5 & 60.5 & 38.0 & 62.8 \\
  DeepSeek-R1 SFT    & 12.4 & 15.6 & 12.3 & 12.4 \\
  DeepSeek-R1 ORL    & 14.7 & 15.9 & 13.2 & 13.3 \\

  \midrule

  Llama-3.2-3B Base & 46.0 & 34.4 & 29.3 & 34.8 \\
  Llama-3.2-3B SFT    & 13.1 & 15.8 & 13.0 & 13.0 \\
  Llama-3.2-3B ORL    & 15.0 & 16.1 & 14.0 & 13.9 \\

  \midrule

    \multicolumn{5}{c}{\textit{Response Length (Llama 3 Tokens)}} \\

  Real                & 16.9 & 20.1 & 38.9 & 18.9 \\

  \midrule

  Llama-3.1-8B Prompt & 39.5 & 49.1 & 53.0 & 42.9 \\
  Llama-3.1-8B SFT    & 11.2 & 12.2 & 29.1 & 11.6 \\
  Llama-3.1-8B ORL    & 11.6 & 16.6 & 35.1 & 12.6 \\

  \midrule

  DeepSeek-R1 Prompt & 414.5 & 424.6 & 472.1 & 416.4 \\
  DeepSeek-R1 SFT    & 12.9  & 13.9  & 37.6  & 14.0 \\
  DeepSeek-R1 ORL    & 11.2  & 13.4  & 20.3  & 12.1 \\

  \midrule

  Llama-3.2-3B Prompt & 52.6 & 61.6 & 61.6 & 58.0 \\
  Llama-3.2-3B SFT    & 11.1 & 12.4 & 31.6 & 11.9 \\
  Llama-3.2-3B ORL    & 11.8 & 15.9 & 29.2 & 13.0 \\

  \midrule

    \multicolumn{5}{c}{\textit{Prompt Length (Llama 3 Tokens)}} \\

  \midrule
    
  Prompted Models & 540.4 & 648.9 & 626.9 & 632.4 \\
  SFT and ORL & 180.4 & 288.9 & 266.9 & 272.4 \\

  \bottomrule

  \end{tabular}
  \caption{Conformity results (Ac = Academic, Jo = Journalistic, Ju = Judicial, Or = Oral).}\label{tab:conformity_results}
  \vspace{-15pt}
\end{table}

\subsection{Evaluation}
\label{sec:evaluation_results}

For evaluation, each six-turn segment is truncated by removing the final elicitor turn, and the model is tasked with generating that turn given the preceding five turns. This is repeated for all segments in the \textit{testing} set, which models have not seen yet. Sample outputs are demonstrated in Appendix~\ref{appendix:sample_responses}.

\subsubsection{Conformity}
\label{sec:conformity_results}

Table~\ref{tab:conformity_results} shows a clear gap between prompt-only models and those adapted to information elicitation via SFT or ORL (e.g., the prompted Llama-3.1-8B averaged a micro perplexity of 46.9 in Academic Interviews, while the ORL model averaged 12.5). This effect is especially strong for DeepSeek-R1, since its reasoning baseline prefaces responses with meta-level deliberation. Fine-tuning suppresses this pattern and brings its outputs in line with human data. Among adapted models, SFT achieves the lowest perplexity, while ORL is slightly higher, consistent with advantage-weighted training prioritizing longer-horizon behavior over token-level likelihood.

In terms of response length, prompted models consistently produce elicitor turns far longer than those observed in real dialogues (e.g., the prompted Llama-3.1-8B averaged 39.5 tokens in Academic Interviews, while the ORL model averaged 11.6 tokens). Beyond misalignment, prompting is also expensive, as it relies on substantially longer prompts (system instructions and dialogue context) and higher token usage (e.g., the prompted models averaged 540.4 tokens in Academic Interviews, while SFT and ORL models averaged 180.4 tokens). Although answer length could be constrained through even longer prompting, doing so is inherently unreliable because it requires the model to deviate from its default behavior. In contrast, SFT and ORL produce elicitor turns much closer to real elicitation behavior with far shorter prompts.

\begin{table}[t!]

  \small

  \centering

  \begin{tabular}{
    m{2.8cm}
    >{\centering\arraybackslash}m{0.7cm}
    >{\centering\arraybackslash}m{0.7cm}
    >{\centering\arraybackslash}m{0.7cm}
    >{\centering\arraybackslash}m{0.7cm}
  }
  
  \toprule

   Model & Ac	& Jo	& Ju	& Or	 \\

  \midrule

    \multicolumn{5}{c}{\textit{Progression}} \\

  Real     & 0.713 & 0.751 & 0.744 & 0.755 \\
  Shuffled & 0.824 & 0.848 & 0.800 & 0.821 \\

  \midrule

  Llama-3.1-8B Prompt & 0.660 & 0.629 & 0.663 & 0.653 \\
  Llama-3.1-8B SFT    & 0.688 & 0.776 & 0.732 & 0.743 \\
  Llama-3.1-8B ORL    & 0.680 & 0.734 & 0.725 & 0.734 \\

  \midrule

  DeepSeek-R1 Prompt & 0.671 & 0.643 & 0.685 & 0.714 \\
  DeepSeek-R1 SFT    & 0.701 & 0.755 & 0.736 & 0.741 \\
  DeepSeek-R1 ORL    & 0.697 & 0.743 & 0.752 & 0.732 \\

  \midrule

  Llama-3.2-3B Prompt & 0.669 & 0.632 & 0.669 & 0.667 \\
  Llama-3.2-3B SFT    & 0.676 & 0.731 & 0.725 & 0.735 \\
  Llama-3.2-3B ORL    & 0.688 & 0.732 & 0.745 & 0.732 \\

  \bottomrule

  \end{tabular}
  \caption{Progression results (Ac = Academic, Jo = Journalistic, Ju = Judicial, Or = Oral). \textit{Real} and \textit{Shuffled} denote human dataset utterances.}\label{tab:main_results}
\end{table}

\begin{table}[t!]

  \small

  \centering

  \begin{tabular}{
    m{2.8cm}
    >{\centering\arraybackslash}m{0.7cm}
    >{\centering\arraybackslash}m{0.7cm}
    >{\centering\arraybackslash}m{0.7cm}
    >{\centering\arraybackslash}m{0.7cm}
  }
  
  \toprule

   Model & Ac	& Jo	& Ju	& Or	 \\

  \midrule

    \multicolumn{5}{c}{\textit{Turn-length Ratio}} \\

  Real                & 1.672 & 3.051 & 1.227 & 3.187 \\  
  Shuffled            & 0.969 & 1.002 & 0.883 & 0.954 \\

  \midrule

  Llama-3.1-8B Prompt & 1.160 & 2.071 & 1.088 & 2.222 \\
  Llama-3.1-8B SFT    & 1.878 & 3.496 & 1.346 & 3.671 \\
  Llama-3.1-8B ORL    & 1.864 & 3.231 & 1.270 & 3.595 \\

  \midrule

  DeepSeek-R1 Prompt & 0.191 & 0.402 & 0.249 & 0.389 \\
  DeepSeek-R1 SFT    & 1.813 & 3.392 & 1.241 & 3.499 \\
  DeepSeek-R1 ORL    & 1.877 & 3.421 & 1.476 & 3.632 \\

  \midrule

  Llama-3.2-3B Prompt & 0.986 & 1.820 & 1.018 & 1.866 \\
  Llama-3.2-3B SFT    & 1.882 & 3.487 & 1.314 & 3.647 \\
  Llama-3.2-3B ORL    & 1.856 & 3.274 & 1.345 & 3.571 \\

  \bottomrule

  \end{tabular}
  \caption{Turn-length ratio results (Ac = Academic, Jo = Journalistic, Ju = Judicial, Or = Oral). \textit{Real} and \textit{Shuffled} denote human dataset utterances.}\label{tab:turn_length_ratio_results}
  \vspace{-10pt}
\end{table}

\subsubsection{Progression} 

Given that there are 6 turns, we compute Progression with $k = 5$, and since only the last utterance is generated we set $\gamma = 0.5$ to more heavily decrease the importance of non-generated utterances. We include as a comparison baseline a version of the test set where turns within a dialogue have been randomly shuffled. Progression embeddings are obtained using the all-MiniLM-L12-v2 model from Sentence Transformers \cite{sentence_transformers_all-minilm-l12-v2_2021}. The real and randomly shuffled references in Table \ref{tab:main_results} show that the Progression metric captures the intended semantic drift within the dialogues. Real human elicitation dialogues consistently fall within a lower range (0.713–0.755), while randomly shuffled dialogues yield much higher values (0.800–0.848). Within this reference frame, both SFT and ORL models closely track real dialogues across domains, whereas prompt-only models are uniformly lower. This pattern indicates that prompted models tend to linger on previously covered topics.

\vspace{-5pt}
\subsubsection{Turn-Length Ratio} 

Table~\ref{tab:turn_length_ratio_results} presents turn-length ratio results, with the shuffled dataset also included. In the shuffled baseline, elicitor and respondent turns are randomly reassigned, causing answers to appear as questions and vice versa. This role mixing neutralizes the systematic asymmetry present in real dialogues, leading to ratios close to 1. Meanwhile, real dialogues show ratios well above 1 across domains, reflecting the elicitation dynamic in which elicitor turns prompt longer respondent utterances. Both SFT and ORL closely reproduce this asymmetry, learning to limit elicitor verbosity in proportions similar to human interviewers, a behavior that prompted models largely fail to exhibit. As with response length, this gap cannot be reliably closed through additional prompting, since doing so does not make models internalize elicitor behavior.

\subsection{Human Evaluation} 

\begin{table}[t!]
    \small
    \centering

    \begin{tabular}{
      >{\centering\arraybackslash}m{1.cm}
      >{\centering\arraybackslash}m{0.8cm}
      >{\centering\arraybackslash}m{0.8cm}
      >{\centering\arraybackslash}m{0.8cm}
      >{\centering\arraybackslash}m{0.8cm}
      >{\centering\arraybackslash}m{0.8cm}
      >{\centering\arraybackslash}m{0.8cm}
    }
    
    \toprule
      
    & P & CC & OR & PE & C \\
    \midrule
    Real     & 3.89 & 3.75 & 3.92 & 3.71 & 3.97 \\
    Prompted & 4.25 & 4.07 & 4.25 & 4.17 & 4.20 \\
    SFT      & 3.65 & 3.64 & 4.01 & 3.57 & 3.96 \\
    ORL      & 3.70 & 3.67 & 3.89 & 3.45 & 3.99 \\

  \bottomrule
  \end{tabular}
  \caption{Human evaluation results per response source. Metrics: P = Progression, CC = Conversational Control, OR = Outcome Relevance, PE = Probing Effectiveness, C = Conformity.}\label{tab:human_eval_per_source}
  \vspace{-15pt}
\end{table}

We collected human ratings of elicitor question quality using Amazon Mechanical Turk. For each HIT, workers reviewed the five-turn dialogue context followed by corresponding real, prompted, SFT, and ORL responses. We randomized the order of the response sources across HITs to prevent order effects. Workers scored questions on five-point scales representing different aspects of elicitation. The full set of instructions given to participants is available in Appendix Table~\ref{appendix:human_eval_part_instructions}. In total, 300 dialogue evaluations were collected (100 dialogue blocks sampled with stratified sampling, rated by three workers). Workers were required to have more than 500 prior HITs and approval rate above 90\%. Workers were compensated at a rate consistent with U.S. minimum wage, based on estimated task completion time.

The results in Table~\ref{tab:human_eval_per_source} show that fine-tuned models more closely approximate the distribution of real utterances than prompted models, supporting automatic evaluation findings. Fine-tuned model scores track the real responses more consistently, even though their absolute ratings are lower. This outcome is particularly interesting, as it suggests that crowd workers rewarded fuller elicitor utterances, which prompted models reliably generate. However, this behavior departs from real dialogue as prompted responses tend to be more verbose, leading, and polished as shown in Section~\ref{sec:evaluation_results}. By contrast, fine-tuned models, while receiving lower absolute scores, overwhelmingly align more closely with the distribution of real utterances.

\subsubsection{Potential Misuse and Mitigations} 

Information elicitation agents raise ethical concerns due to their goal of guiding conversations toward information that serves external objectives, which can introduce risks of biased or leading questioning, power imbalances, and undue influence, particularly in domains such as journalism, legal proceedings, or public opinion polling where neutrality is critical. These risks are compounded in settings involving sensitive personal information (e.g., healthcare or counseling), where issues of data privacy, storage, and potential misuse arise. Mitigations include enforcing constraints on questioning strategies (e.g., prompt-level guidelines or post-generation filtering) and extending evaluation protocols to detect and penalize biased behavior. Systems should also implement data minimization, secure storage, and clear user consent and transparency practices. In high-stakes or sensitive contexts, IEAs should only be deployed under appropriate institutional and regulatory oversight (e.g., IRB or clinical governance), with human monitoring and auditing to ensure compliance with ethical and legal standards, rather than relying solely on model-level safeguards.

\section{Conclusion}
\label{sec:conclusion}

Our evaluation demonstrates YIELD effectively supports the development of models for the elicitation setting and our finite-horizon POMDP formulation with offline reinforcement learning shows that dialogue-level optimization is competitive with SFT even when trained with a minimal reward signal. This work opens several directions for future research including extended reward formulations that capture domain-specific notions of informational value, interactive evaluation with live respondents, and additional evaluation dimensions that capture a broader range of elicitation characteristics.

\section{Limitations}
\label{sec:limitations}

The reward formulation used in offline reinforcement learning is intentionally minimal and may not adequately capture informational value across elicitation contexts. Additionally, evaluation is limited to offline, next-turn generation on fixed contexts rather than interactive settings. Moreover, some proposed metrics are sensitive to design choices (e.g., Progression depends on the choice of embedding model and hyper-parameters) and may not capture all aspects relevant to effective elicitation behavior. Finally, this work focuses on modeling the base elicitor agent and does not address conversation management. While a fine-tuned LLM can serve as the core of an elicitation system, it is likely insufficient on its own for sustaining structured and effective long-horizon interactions.

\section*{Ethical Considerations}

We have taken all reasonable steps to ensure that the content in this dataset either resides in the public domain or adheres to applicable licenses, such as Creative Commons. We encourage users to notify us of any content that may be subject to restrictions not accounted for so that we can address the matter appropriately and update the dataset as necessary.


\section*{Acknowledgments}

This research was supported by U.S. National Science Foundation grant number IIS-2336768. Any opinions, findings, conclusions, or recommendations expressed in this paper are of the authors and do not necessarily reflect those of the sponsor.


\bibliography{references/references}

\appendix

\renewcommand{\thefigure}{\Alph{section}.\arabic{figure}}
\renewcommand{\thetable}{\Alph{section}.\arabic{table}}
\setcounter{figure}{0}
\setcounter{table}{0}

\section{Appendix}

\subsection{Data Sources}
\label{appendix:data_sources}

\subsubsection{Oral History}

We obtained documents for the Oral History domain from a variety of government sources. The National Archives and Records Administration (NARA) is the U.S. government agency responsible for preserving and providing access to federal records and historical documents. We obtained the following from this source: National Archives History Office Interviews \cite{the_us_national_archives_and_records_administration_national_2025}, National Personnel Records Center Oral History Interviews \cite{the_us_national_archives_and_records_administration_national_2023}, Veterans Oral History Interviews \cite{the_us_national_archives_and_records_administration_veterans_2024}, and National Archives Assembly's Legacy Project \cite{the_us_national_archives_and_records_administration_national_2024}. NARA manages the John F. Kennedy Presidential Library, which offers collections related to John F. Kennedy's life, career, and era, along with his administration's work and its influence on the public. We added the Returned Peace Corps Volunteer Collection \cite{the_john_f_kennedy_library_returned_2025} to YIELD from this source. Lastly, we included the NASA Johnson Space Center Oral History Project \cite{national_aeronautics_and_space_administration_-_johnson_space_center_oral_2021}, which documents the personal experiences of individuals who played vital roles in America’s space programs. Participants include astronauts, engineers, and administrators.


\subsubsection{Judicial Proceedings}

We obtained the Supreme Court case collection for YIELD from the Oyez repository developed by Cornell's Legal Information Institute, Justia, and Chicago-Kent College of Law. The Supreme Court hears cases from the first Monday in October until late June or early July (a ``term'') \cite{supreme_court_of_the_united_states_court_2025}. Hence, the ``years'' included in the dataset start in October. As of this writing, the last complete term is 2023. While the archive runs from 1955, we only include YIELD in the last 10-year period (2014-2023) to make the dialogue pool in this domain not be overwhelmingly larger than other domains.


\subsubsection{Academic Interviews}

Most collections for the Academic Interviews category originated from the Harvard Dataverse Repository, an open-access platform for researchers across disciplines to share research data. We obtained the following collections from this source: Covid Threshold Project Dataverse \cite{lachman_interview_2022}, Healthworker Interview Transcripts \cite{watkins_healthworker_2021}, Drivers of Food Choice \cite{girard_dfc_tanzania_2019}, AI feedback: Moving beyond the hype \cite{tay_replication_2024,tay_ai_2024}, Relationship Building Around Farmers’ Market Incentive Programs (D.C. Market Manager Relationships) \cite{policastro_relationship_2017}, Leaders Leading Organizational Change \cite{sadaric_interviews_2022}, and Biodiversity Offsetting \cite{taherzadeh_interview_2016}. We additionally obtained documents from the Flu Vaccination Interviews \cite{price_barriers_2022,price_interview_2021} for this domain.


\subsubsection{Journalistic Investigations}

We sourced collections for the Journalistic Investigations domain from Voice of America (VOA) \cite{voice_of_america_voanewscom_nodate} and Wikinews \cite{wikinews_categoryinterviews_nodate}. VOA is a U.S. international broadcaster under the U.S. Agency for Global Media. Since VOA is funded by Congress, all VOA content is in the public domain, including the ``VOA Interview'' series which we used for YIELD. Wikinews is a collaborative news platform that the Wikimedia Foundation runs. It publishes user-generated reports in multiple languages under a Creative Commons license. We selected a subset of articles from the ``Interviews'' category.

\subsection{Experimental Dataset Statistics}

Table~\ref{tab:experimental_dataset_statistics} shows statistics of the experimental dataset used for SFT and ORL.

\subsection{Licensing}
\label{appendix:licensing}

Table~\ref{tab:licensing} provides a summary of copyright designations by collection.

\begin{table}
    \centering
    \small
    \begin{tabular}{|m{1.5cm}|m{5.3cm}|}
        \hline
        \textbf{Copyright} & \textbf{Collections} \\ \hline
  
          CC0 1.0 & Covid Threshold Project Dataverse, Healthworker Interview Transcripts, Drivers of Food Choice, AI feedback: Moving beyond the hype, Relationship Building Around Farmers’ Market Incentive Programs (D.C. Market Manager Relationships), Leaders Leading Organizational Change, Biodiversity Offsetting \\ \hline
          
          CC BY 4.0 & Flu Vaccination Interviews, Wikinews \\ \hline
          
          CC BY-NC 4.0 & Oyez Supreme Court \\ \hline
    
          Public Domain & National Archives History Office Interviews, National Personnel Records Center Oral History Interviews, Veterans Oral History Interviews, National Archives Assembly's Legacy Project, Returned Peace Corps Volunteer Collection, VOA News \\ \hline
          
    \end{tabular}
    \captionsetup{aboveskip=0pt, belowskip=0pt}
    \caption{Collections by copyright status.}
    \label{tab:licensing}
  \end{table}

\subsection{Experimental Dataset Sample}
\label{appendix:experimental_dataset_sample}

The following sample shows a line from the experimental dataset.

\begin{quote}
\small

    "\textbf{block\_id}": "voa-news-00034:11", 
    
    "\textbf{domain}": "journalistic\_investigations", 
    
    "\textbf{factual\_novelty\_score}": 2, 
    
    "\textbf{return\_to\_go}": 18.550635451980344, 
    
    "\textbf{messages}": [
    
    \{"\textit{role}": "system", "\textit{content}": "Act as an information elicitation agent for journalistic investigations."\}, 
    
    \{"\textit{role}": "user", "\textit{content}": "That's correct, exactly."\}, 
    
    \{"\textit{role}": "assistant", "\textit{content}": "The decider would be the president who was elected."\}, 
    
    \{"\textit{role}": "user", "\textit{content}": "Precisely."\}, 
    
    \{"\textit{role}": "assistant", "\textit{content}": "When did you become the national security adviser?"\}, 
    
    \{"\textit{role}": "user", "\textit{content}": "In April of 2018, and I lasted seven, 17 months until September of 2019."\}, 
    
    \{"\textit{role}": "assistant", "\textit{content}": "You were not the first national security adviser under this president?"\}
    
    ]\}

\end{quote}

\subsection{Baseline's Prompt}
\label{appendix:baseline_prompt}

The following is the prompt used on the baseline models. The prompt was not tailored to any specific model and was generated using an LLM to avoid researcher bias. 

\begin{quote}
\small

You are participating in an oral history interview 
as the "elicitor" (interviewer). Your role is to 
elicit personal stories, reflections, and details 
from the "respondent" (interviewee) through natural, 
conversational questions. 

Follow these principles:

\begin{itemize}
    \item Be friendly, engaged, and curious.
    \item Ask clear, open-ended questions.
    \item Let the conversation flow naturally build on what the provider just said.
    \item Use informal, real spoken language (e.g., “So tell me…” or “What was that like?”).
    \item Occasionally use acknowledgments or restatements like “OK” or “Wow” to maintain flow.
    \item Focus on memories, motivations, feelings, and personal experiences.
\end{itemize}

Here’s an example start:

[conversation]

elicitor: OK, Larry, why don’t you tell us a little 
bit about what you're doing now.\\
respondent: I'm retired and I do a lot of scroll 
signs, which keeps me busy in the cold months of 
the winter and the hot months of the summer.\\
elicitor: Where are you living now?\\
respondent: In Green Bay, Wisconsin. We farmed for 
30 years and then I retired and we bought a farm 
house and remodeled that...\\
elicitor: Too much work. OK.\\
respondent: So that’s brought me up to today.\\
elicitor: Well, then the next thing that I want to 
do is have you think back to the year that you 
joined the Peace Corps. What were you doing before? 
How did you hear about it? What made you decide to 
join?\\
respondent: [responds...]\\
elicitor: [Your turn — ask the next natural question 
based on what they just said.]

\end{quote}

\subsection{Instructions Given to Human Evaluation Participants}
\label{appendix:human_eval_part_instructions}

The human evaluation criteria provided to workers consists of Progression (P), Conversational Control (CC), Outcome Relevance (OR), Probing Effectiveness (PE), and Conformity (C). The following is the full set of instructions given to human evaluation participants:

\begin{quote}
\small

\textbf{Your Task}

You will read short conversation snippets between an interviewer (denoted \textbf{Seeker}) and an interviewee (denoted \textbf{Provider}).

\begin{itemize}
    \item Each snippet has 5 turns of context (the conversation so far).
    \item After the context, you will see 4 distinct interviewer (\textbf{Elicitor}) questions.
    \item Your job is to rate that \textbf{Elicitor} question on the five Metrics listed below.
    \item A Dialogue Setting is provided to put the conversation in context.
\end{itemize}

\textbf{Metrics:}

\begin{itemize}
\item \textbf{Conformity (1–5)}\\
How well does the question match the expected style of the speaker, per the Dialogue Setting?\\
1 = Very unnatural / out of character\\
3 = Neutral / acceptable but not strong\\
5 = Perfectly fits the expected style\\

\item \textbf{Progression (1–5)}\\
Does the question move the conversation forward?\\
1 = Off-topic or repetitive\\
3 = Somewhat helps the discussion continue\\
5 = Strongly advances the discussion\\

\item \textbf{Outcome Relevance (1–5)}\\
Is the question relevant to the goal of the exchange, per the Dialogue Setting?\\
1 = Unrelated to the goal\\
3 = Moderately related\\
5 = Directly addresses the goal\\

\item \textbf{Probing (1–5)}\\
Does the question push for deeper information?\\
1 = No probing at all\\
3 = Some attempt to get more detail\\
5 = Strong, pointed probing\\

\item \textbf{Control (1–5)}\\
Does the question guide or steer the conversation?\\
1 = No control, just follows along\\
3 = Some influence\\
5 = Clearly sets direction\\
\end{itemize}

After you finish rating all the responses, click \textbf{Next HIT} to move on.

\textbf{Dialogue Setting}

\${domain}

\textbf{Dialogue Context}

\${context\_turns}

\end{quote}

\subsection{Sample Responses}
\label{appendix:sample_responses}

The following shows sample responses from Llama-3.1-8B models.

\begin{quote}
\small

    \textbf{Block ID}: "jsc-oral-history-00206:183"

    \textbf{Input Prompt}: "Act as an information elicitation agent for oral history. \textit{Respondent}: Yes. And I feel I was very fortunate to be where I was, when I was. It was a different situation that is unique to where I was. Other people were part of the program, but somehow or other, I felt like I was in a unique situation through most of it. \textit{Elicitor}: It certainly was a very unique time. \textit{Respondent}: Yes. \textit{Elicitor}: To be working in and living through. Looking back over your time at NASA, was there any point that you would consider—or any two points that might be your most challenging moments? And then maybe that you feel is your most significant accomplishment? \textit{Respondent}: Challenging moments, I guess revolved around the guys like Charlie Bassett, you know. I was reasonably close. And Gus [Virgil I. Grissom]. That was a challenging time. That was a tough time. As far as the years, I think those kind of went along with the way things ended up. The times that there was a big smile on your face, I think were part of the end of the success as to how things worked out...

    \textbf{Responses}

    \underline{Real}: "Well, things certainly did work out very well."

    \underline{Prompted}: "It sounds like those were really difficult times, losing friends and colleagues like that. Gus Grissom, of course, was a very important part of the Gemini program. And Charlie Bassett was a great guy, too. You mentioned the end of the program, and that time frame - the end of the success, I think you said. What was it like, feeling like you'd accomplished what you set out to do, and then seeing the program come to a close?" 
    
    \underline{SFT}: "And you mentioned that you'd like to talk about some of the people that you worked with."
    
    \underline{ORL}: "Was there a time when you realized that the Space Program was really taking off?"

\end{quote}

\end{document}